\documentclass{article}

\usepackage[final]{neurips_data_2022}

\usepackage[utf8]{inputenc} 
\usepackage[T1]{fontenc}    
\usepackage{hyperref}       
\usepackage{url}            
\usepackage{booktabs}       
\usepackage{amsfonts}       
\usepackage{nicefrac}       
\usepackage{microtype}      
\usepackage{xcolor}         
\usepackage{graphics}
\usepackage{wrapfig}

\usepackage{graphicx}
\usepackage[linesnumbered,ruled,noend]{algorithm2e}
\usepackage{algpseudocode}%
\usepackage{balance} 
\usepackage{todonotes}
\usepackage{color}
\usepackage{colortbl}
\usepackage{comment}
\usepackage{xurl}
\usepackage{soul}
\usepackage{multirow}
\usepackage{caption}
\usepackage{subcaption}
\usepackage{multicol}
\usepackage{bigstrut}
\usepackage[justification=centering]{caption}
\usepackage{threeparttable}
\title{METS-CoV: A Dataset of Medical Entity and Targeted Sentiment on COVID-19 Related Tweets}

\author{%
  Peilin Zhou$^{1,2}$ \quad
  Zeqiang Wang$^{1,2}$  \quad
  Dading Chong$^{3}$ \quad
  Zhijiang Guo$^{4}$ 
  \\
  \textbf{Yining Hua}$^{5}$ \quad
  \textbf{Zichang Su}$^{1,6}$ \quad
   \textbf{Zhiyang Teng}$^{7}$ \quad
   \textbf{Jiageng Wu}$^{1,2}$ \quad
  \textbf{Jie Yang}$^{1,2}$\thanks{Corresponding Author.}\vspace{1mm}\\
$^{1}$School of Public Health and the Second Affiliated Hospital, Zhejiang University, China\vspace{1mm}\\
$^{2}$The Key Laboratory of Intelligent Preventive Medicine of Zhejiang Province, China\vspace{1mm}\\
$^{3}$School of Electronic and Computer Engineering, Peking University, China\vspace{1mm}\\
$^{4}$Department of Computer Science and Technology, University of Cambridge, UK\vspace{1mm}\\
$^{5}$Department of Biomedical Informatics, Harvard Medical School, USA\vspace{1mm}\\
$^{6}$Chu Kochen Honors College, Zhejiang University, China\vspace{1mm}\\
$^{7}$School of Engineering, Westlake University, China\vspace{1mm}\\
  \texttt{\{zhoupalin, jieynlp\}@gmail.com, \{wzq99, suzc,jiagengwu\}@zju.edu.cn} \\
  \texttt{ 1601213984@pku.edu.cn, tengzhiyang@westlake.edu.cn} \\
  \texttt{zg283@cam.ac.uk, yining\_hua@hms.harvard.edu} \\
}

\begin{document}

\maketitle

\begin{abstract}
The COVID-19 pandemic continues to bring up various topics discussed or debated on social media. In order to explore the impact of pandemics on people's lives, it is crucial to understand the public's concerns and attitudes towards pandemic-related entities (e.g., drugs, vaccines) on social media. However, models trained on existing named entity recognition (NER) or targeted sentiment analysis (TSA) datasets have limited ability to understand COVID-19-related social media texts because these datasets are not designed or annotated from a medical perspective. This paper releases METS-CoV, a dataset containing medical entities and targeted sentiments from COVID-19-related tweets. METS-CoV contains 10,000 tweets with 7 types of entities, including 4 medical entity types (\textit{Disease}, \textit{Drug}, \textit{Symptom}, and \textit{Vaccine}) and 3 general entity types (\textit{Person}, \textit{Location}, and \textit{Organization}). To further investigate tweet users' attitudes toward specific entities, 4 types of entities (\textit{Person}, \textit{Organization}, \textit{Drug}, and \textit{Vaccine}) are selected and annotated with user sentiments, resulting in a targeted sentiment dataset with 9,101 entities (in 5,278 tweets). To the best of our knowledge, METS-CoV is the first dataset to collect medical entities and corresponding sentiments of COVID-19-related tweets. We benchmark the performance of classical machine learning models and state-of-the-art deep learning models on NER and TSA tasks with extensive experiments. Results show that the dataset has vast room for improvement for both NER and TSA tasks. With rich annotations and comprehensive benchmark results, we believe METS-CoV is a fundamental resource for building better medical social media understanding tools and facilitating computational social science research, especially on epidemiological topics. Our data, annotation guidelines, benchmark models, and source code are publicly available (\url{https://github.com/YLab-Open/METS-CoV}) to ensure reproducibility. 
\end{abstract}

\section{Introduction}
\begin{figure}[htbp]
\centering
\centering
\includegraphics[width=110mm]{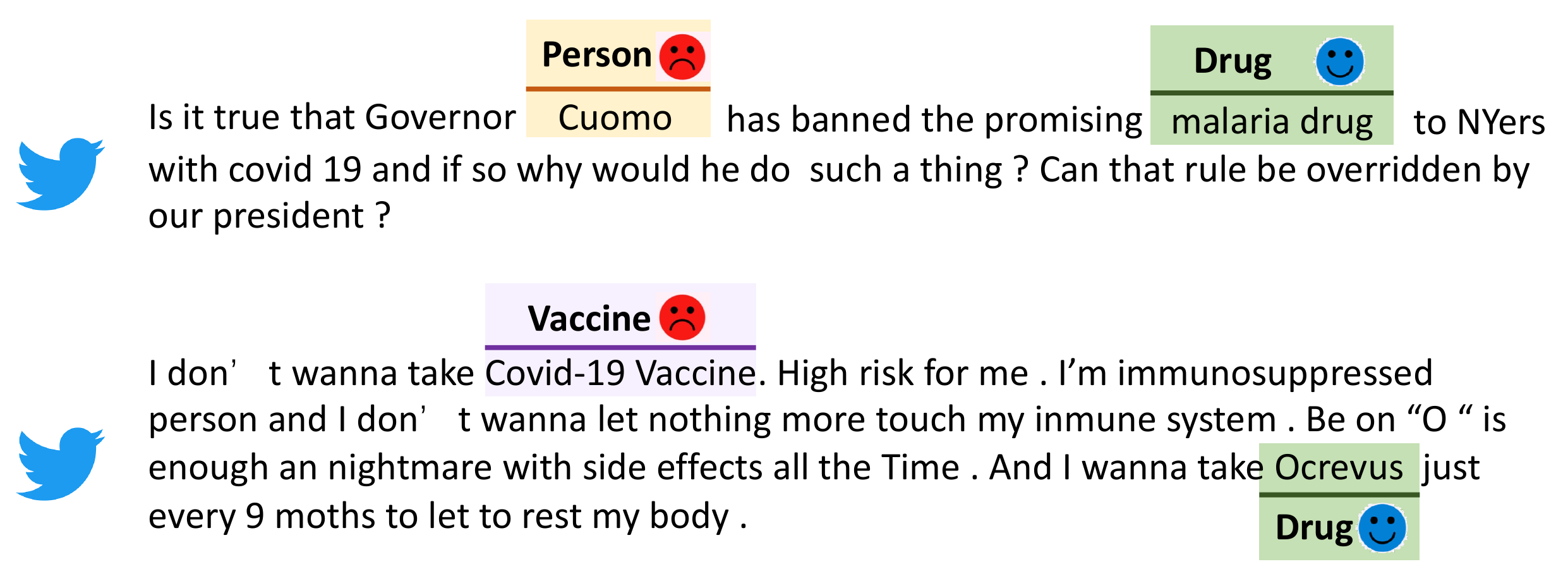}
\caption{Examples of medical entities and targeted sentiments in tweets.}
\label{Figure:data_examples}
\end{figure}

The outbreak of the COVID-19 pandemic has had severe implications for people's lives and global health~\citep{covid-19-pandemic}. To assess the impact of this pandemic on the public, epidemiologists and medical researchers conducted research through various methods such as clinical follow-up~\citep{follow-up}, questionnaires~\citep{questionare}, and app tracking~\citep{app_tracking}. Social media is one of the most popular media types in the world. With its large user base, rapid information dissemination, and active participation, social media has become an important channel for the public to express their feelings and opinions on COVID-19-related topics, which provides researchers with large-scale and low-cost materials to track the impact of COVID-19. For example, as shown in Fig.\ref{Figure:data_examples}, Twitter users express their attitudes toward medical entities (e.g., drugs and vaccines) through their Twitter posts. Large-scale analyses of such tweets are vital to understanding the public's opinions towards medical topics, which can further help medical research or public health management. Therefore, many studies have been proposed to track and analyze people through social media platforms such as Twitter, one of the most widely used social media platforms~\citep{study1,study2,study3,study4,study5}.

However, existing natural language processing (NLP) tools~\citep{nltk,tools_compare,mengcao} struggle to fulfill the surging demand for accurate COVID-19 tweet analysis due to several reasons: 
\begin{itemize}
    \item Most NLP tools are not explicitly designed for social media texts, leading to dramatic performance degradation when applied to noisy texts such as tweets~\citep{limits_nlp_tool}.
    \item Most current NLP tools are developed for general applications, so domain adaptation can be poor without adding external knowledge from medical studies.
    \item Existing NLP tools are not designed from the medical or public health research perspective, making it challenging to meet the needs of epidemiologists to analyze medical topics.
\end{itemize}
In short, the very underlying reason for the poor applicability of these tools in COVID-19 social media studies is the lack of COVID-19-related social media datasets designed and annotated from a medical perspective. Furthermore, one of the most critical analysis goals for COVID-19-related studies conducted on tweets is to find the entities (both general and medical entities) discussed by users and their attitudes or feelings towards them. Such goals correspond to two basic NLP tasks: named entity recognition (NER) and targeted sentiment analysis (TSA). NER aims at entity extraction from unstructured texts, and TSA aims at user sentiment prediction towards the targeted entities.

In this work, we release \textbf{METS-CoV} (\textbf{M}edical \textbf{E}ntities and \textbf{T}argeted \textbf{S}entiments on \textbf{CoV}id-19-related tweets), a dataset that contains 10,000 tweets annotated with 7 types of entities, including 4 medical entity types (\textit{Disease}, \textit{Drug}, \textit{Symptom}, and \textit{Vaccine}) and 3 general entity types (\textit{Person}, \textit{Location}, and \textit{Organization}). In addition, 4 types of entities (\textit{Person}, \textit{Organization}, \textit{Drug}, and \textit{Vaccine}) are selected and annotated with user sentiments to explore the attitudes of tweet users toward specific entities. Unlike other general NER and TSA datasets, METS-CoV is built from a public health research perspective and contributes to developing tailored natural language processing tools for the medical domain to mine valuable information from social media rather than only relying on keyword search.
For example, based on the NER and TSA model trained on METS-CoV dataset, researchers can track public attitudes toward COVID-19 vaccination for more efficient vaccination policies. The models could also track the public’s mental status in different COVID-phases, providing potential solutions for addressing the global mental health crisis.
In addition, we design detailed annotation guidelines for medical entities on tweets. The guidelines have been applied to our annotation process as strict supervision and control to ensure quality. We benchmark the performance of classical machine learning models and state-of-the-art deep learning models (including pretraining language models) on NER and TSA tasks of METS-CoV. According to the benchmark results, the NER and TSA tasks on the dataset have much space for performance improvement.

\section{Related Work}
METS-CoV supports two basic NLP tasks from a medical perspective: 1) named entity recognition, i.e., identifying general and medical entities, and 2) targeted sentiment analysis, i.e., predicting the attitudes of Twitter users towards specific entities (including \textit{Drug} and \textit{Vaccine}). This section reviews several commonly used open-source datasets for these two tasks and compares them with the proposed dataset.

\subsection{Named Entity Recognition Datasets}
CoNLL 2003~\citep{tjong-kim-sang-de-meulder-2003-introduction} is one of the most widely used NER datasets with its newswire texts collected from the Reuters Corpus. The dataset consists of 4 general entity types: PER (Person), LOC (Location), ORG (Organization), and MISC (Miscellaneous), which are also adopted in the SciTech News dataset~\citep{jia-etal-2019-cross}. 
WNUT NER~\citep{strauss-etal-2016-results} is a benchmark NER dataset for the social media domain that consists of manually annotated tweets with 10 entity types. Nevertheless, none of the entity types is medical-related. Similarly, the recently release Tweebank-NER dataset is neither medical-related \citep{jiang-EtAl:2022:LREC2}.
In the medical domain, NER is often used to extract medical terminologies from clinical case reports (CCRs) or electronic medical records (EMRs). A representative medical NER dataset is i2b2-2010 dataset~\citep{uzuner20112010}, which includes discharge summaries and progress notes provided by well-known medical centers with 3 entity types: test, problem, and treatment. Besides, one of the SMM4H shared tasks \citep{klein-etal-2020-overview,weissenbacher-etal-2019-overview} released a dataset for extracting tweet text spans with adverse drug reactions (ADR). This dataset is not COVID-19-related.
On the other hand, the CORD-NER dataset~\citep{wang2020comprehensive} has 75 fine-grained types of entities from scientific papers about COVID-19 and historical coronavirus research. But since social media texts have way more colloquial forms than scientific papers, models trained on WNUT or CORD-NER are unsuitable for social media analyses.
\subsection{Targeted Sentiment Analysis Datasets}
Most TSA studies typically experiment on 3 datasets: LAPTOP~\citep{pontiki-etal-2014-semeval}, TWITTER~\citep{mitchell-etal-2013-open}, and REST~\citep{pontiki-etal-2015-semeval,pontiki-etal-2016-semeval}. Specifically, LAPTOP and REST are user review datasets collected from the laptop and restaurant domains. The TWITTER dataset has tweets but only with general types of entities (\textit{Person} and \textit{Organization}). At the same time, the data might be outdated for the ever-involving social media languages.

There are several recent open-domain TSA datasets. For example, YASO~\citep{orbach-etal-2021-yaso} is an open-domain TSA dataset containing more than 2,000 English user comments extracted from YELP~$\footnote{https://www.yelp.com/dataset}$, AMAZON~\citep{amazon}, SST~\citep{SST}, and OPINOSIS~\citep{opinosis}, covering a variety of topics in multiple domains. COVIDSenti~\citep{naseem2021covidsenti} includes 90,000 COVID-19-related tweets annotated with overall sentiment polarity.

Despite the existing open-domain and in-domain datasets, NER and TSA on clinical social media texts remain an under-explored area. There is a pressing need for such datasets to facilitate social media-based public health studies. To fill in this gap, we release METS-CoV, a COVID-19 tweets-based NER and TSA dataset with 3 general entity labels (\textit{Person}, \textit{Location}, \textit{Organization},) 4 medical entity labels (\textit{Disease}, \textit{Drug}, \textit{Symptom}, and \textit{Vaccine}) as well as sentiment labels for \textit{Person}, \textit{Organization}, \textit{Drug}, and \textit{Vaccine} entities.
    
\section{METS-CoV}

In this section, we provide a detailed description of the collection methodology, annotation process, and statistics of the NER and TSA subsets of METS-CoV.

\subsection{Data Collection Methodology}
We collect COVID-19 related tweets ranging from February 1, 2020, to September 30, 2021, whose unique Tweet Identifier (Tweet ID) came from an open-source database~\citep{Public_Coronavirus_Twitter}. All the tweets are downloaded following Twitter's automation rules and data security policy. For data filtering, we first remove non-English tweets and retweets, resulting in 368,816,761 tweets. Then we filter out the tweets containing URLs because 
they are often restatements of third-party messages and do not directly reflect the users' intentions and attitudes. Finally, we use a list of symptoms (including symptoms of COVID-19 as well as common diseases) as keywords to match the tweets to extract medical-related tweets~\citep{symtom_keywords1,symtom_keywords2,symtom_keywords3,symtom_keywords4,symtom_keywords5}. 2,208,676 tweets remain after the pre-processing step.

\subsection{Data Annotation Process}
We define 7 entity types based on public health research needs \citep{tsao2021social,xu2022novel}, including 3 general entity types and 4 medical entity types for annotation. In particular, we select 4 entity types for additional sentiment annotation with 3 types of sentiment labels: positive, negative, and neutral. All the annotation work is done using the YEDDA annotation platform by~\cite{yang-etal-2018-yedda}. 
We first randomly sample 6,000 tweets from the pre-processed tweets for NER annotation. Then we use these 6,000 annotated NER data to train a BERT-based NER tagger and annotate the rest of the tweets. In order to include more medical entities in the dataset, we select additional 4,000 tweets from the model labeled data (with higher drug and vaccine entity ratios) and manually validate the entities to extend the dataset to a total number of 10,000 tweets.

Here we describe detailed annotation guidelines and processes for the METS-CoV-NER dataset and the METS-CoV-TSA dataset in detail. Note that all our annotators are from medical domains, including medicine, public health and pharmaceutical sciences.

\vspace{3mm}
{\bf The annotation process of METS-CoV-NER}. The annotation includes 3 phases: 

\begin{enumerate}
    \item  In the pre-annotation phase, all annotators are requested to conduct 3 rounds of annotation (with training). F1 value is used as the metric of inter-annotator agreement. All the annotators are assigned the same corpus and required to annotate following the guidelines. After annotation, the project leader compares all the labels and determined the final gold labels, which are used to calculate the inter-annotator agreement. Annotators with F1 greater than 80\% are selected to enter the formal annotation process. The annotation guidelines are also iteratively updated throughout this process.
    \item In the formal annotation phase, annotators label the tweets in pairs (3 pairs in total) to ensure each tweet is annotated twice. When an inconsistency occurs, another annotator steps in to determine the final annotation of the tweet. 
    \item After the formal annotation phase, the project team conducts a quality control check on the labeled results to ensure that the annotated tweets meet the annotation guidelines' requirements. The final inter-annotator agreement is 85.0\% in the F1 value.
\end{enumerate}
{\bf The annotation process of METS-CoV-TSA}.
Similar to the NER dataset's annotation process, METS-CoV-TSA's annotation has 3 phases.
\begin{enumerate}
    \item First, we conduct 6 rounds of sentiment pre-labeling and update the labeling guidelines in each iteration. We use accuracy as the metric of inter-annotator agreement. The procedure to determine the gold labels is the same as for METS-CoV-NER.The annotators who meet the consistency criteria are selected to participate in the subsequent annotation process.
    \item We randomly pair up the annotators (4 pairs in total) and assign the same tweets to each group. A third annotator determines the final annotation when inconsistency occurs.
    \item The project team conducts a secondary check to ensure that the annotation meets the guidelines. The final inter-annotator agreement is 78.4\% in accuracy.
\end{enumerate}

Our annotation guidelines are customized to understand medical-related entities and sentiments better. NER guidelines include rules such as "when the manufacturer name refers to a vaccine in the tweet context, the name should be annotated as a vaccine rather than an organization." Sentiment guidelines include rules such as "when irony appears in an annotation entity, its sentiment should be annotated from the standpoint of the person who posted the tweet."
More details can be found in our guidelines (see supplementary data).
\newpage
\subsection{Dataset Statistics}
\begin{wrapfigure}{r}{0cm}
\centering
\includegraphics[width=70mm]{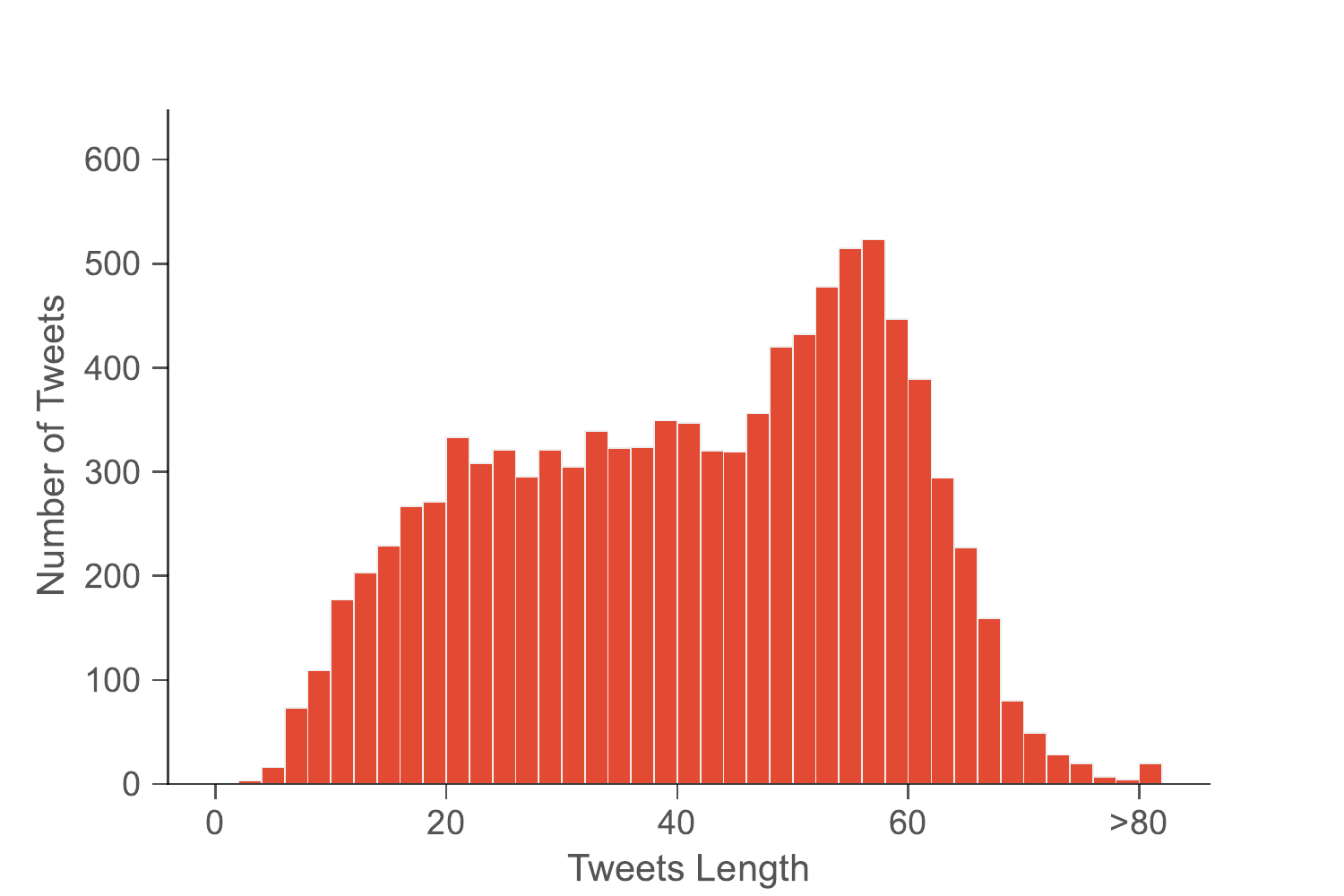}
\caption{The distribution of tweets length of METS-CoV.}
\label{Figure:data_len_distribution}
\end{wrapfigure}
Fig.\ref{Figure:data_len_distribution} shows the distribution of tweet lengths in METS-CoV. Most tweets have lengths shorter than 80 (split with white spaces). Among them, The largest proportion of tweets have a length of around 50. Table~\ref{tab:ner_data} shows the statistics of the METS-CoV-NER dataset. 10,000 tweets contain 19,057 entities in total, resulting in 1.91 entities per tweet in this dataset. We observe that \textit{Symptom} entities have the highest frequency. This is an expected 
result of pre-filtering tweets using symptom-related keywords$\footnote{Some matched symptom keywords are not real symptom entities in tweets due to the ambiguity of words, that is why the total number of Symptom entities is less than 10,000.}$. Other than \textit{symptoms}, all other 6 entity types have
relatively balanced proportions. 
Table~\ref{tab:tsa_data} presents the statistical information of the METS-CoV-TSA dataset, where we observe that neutral sentiment accounts for the highest proportion across all 4 target entities. For the \textit{drug} entities, users have significantly more positive sentiments than negative sentiments, whereas for the \textit{vaccine} entities, users have similar positive and negative sentiments. 
\vspace{5mm}

\begin{minipage}{\textwidth}
\begin{minipage}[t]{0.495\textwidth}
\makeatletter\def\@captype{table}
\centering
\footnotesize 
\renewcommand\arraystretch{0.8}   
\caption{Statistics of METS-CoV-NER dataset.}
\label{tab:ner_data}
    \begin{tabular}{|l|c|c|c|}
    \hline
    \multicolumn{1}{|c|}{\textbf{Number}} & \textbf{Train} & \textbf{Dev}   & \textbf{Test} \bigstrut\\
    \hline
    Tweets & 7,000 & 1,500 & 1,500 \bigstrut\\
    \hline
    Tokens & 285k  & 59k   & 60k \bigstrut\\
    \hline
    All Entities & 13,570 & 2,749 & 2,738 \bigstrut\\
    \hline
    $\;\;\;\;\;$ Person & 2,253 & 472   & 487 \bigstrut\\
    \hline
    $\;\;\;\;\;$ Location & 1,371 & 294   & 279 \bigstrut\\
    \hline
    $\;\;\;\;\;$ Organization & 1,934 & 396   & 381 \bigstrut\\
    \hline
    $\;\;\;\;\;$ Disease & 1,555 & 301   & 258 \bigstrut\\
    \hline
    $\;\;\;\;\;$ Drug  & 1,077 & 262   & 227 \bigstrut\\
    \hline
    $\;\;\;\;\;$ Symptom & 4,223 & 806   & 869 \bigstrut\\
    \hline
    $\;\;\;\;\;$ Vaccine & 1,157 & 218   & 237 \bigstrut\\
    \hline
    \end{tabular}%
\end{minipage}
\begin{minipage}[t]{0.49\textwidth}
\makeatletter\def\@captype{table}
\centering
\footnotesize 
\renewcommand\arraystretch{0.8}  
\caption{Statistics of METS-CoV-TSA dataset.}
\label{tab:tsa_data}

   \begin{tabular}{|l|c|c|c|c|}
    \hline
    \multicolumn{2}{|c|}{\textbf{Number}} & \textbf{Train} & \textbf{Dev} & \textbf{Test} \bigstrut\\
    \hline
    \multirow{3}[6]{*}{Person} & POS   & 260   & 64    & 58 \bigstrut\\
\cline{2-5}          & NEU   & 1293  & 256   & 240 \bigstrut\\
\cline{2-5}          & NEG   & 700   & 152   & 189 \bigstrut\\
    \hline
    \multirow{3}[6]{*}{Organization} & POS   & 126   & 24    & 31 \bigstrut\\
\cline{2-5}          & NEU   & 1346  & 284   & 251 \bigstrut\\
\cline{2-5}          & NEG   & 462   & 88    & 99 \bigstrut\\
    \hline
    \multirow{3}[6]{*}{Drug} & POS   & 234   & 85    & 64 \bigstrut\\
\cline{2-5}          & NEU   & 730   & 147   & 142 \bigstrut\\
\cline{2-5}          & NEG   & 113   & 30    & 21 \bigstrut\\
    \hline
    \multirow{3}[6]{*}{Vaccine} & POS   & 112   & 25    & 20 \bigstrut\\
\cline{2-5}          & NEU   & 913   & 173   & 183 \bigstrut\\
\cline{2-5}          & NEG   & 132   & 20    & 34 \bigstrut\\
    \hline
    \end{tabular}%
\end{minipage}
\end{minipage}
\section{Model Benchmarking}
In this section, we evaluate the performance of statistical machine learning models, neural networks, general domain large-scale pre-trained language models (PLM), and COVID-19-related PLM for the NER task and the TSA task on METS-CoV, respectively. In addition, we select the best model from each group for in-depth analysis and discussion.

\vspace{3mm}
\subsection{Named Entity Recognition}
{\bf Models}. The NER models that we benchmark on the NER dataset can be divided into 4 branches: a traditional statistical machine learning model: Conditional Random Field (CRF)~\citep{lafferty2001conditional}; 6 neural network models from a combination of (1) a BiLSTM for word-level feature extraction
(denoted as WLSTM in Table~\ref{table:ner_benchmark}) and (2) CRF or non-CRF as the inference layer or (3) without character information or encoding character feature with structure CNN, LSTM (denoted as CCNN and CLSTM respectively)~\citep{YangSC17,lample-etal-2016-neural,HuangXY15,ma-hovy-2016-end}; 3 general domain PLM including BERT~\citep{devlin-etal-2019-bert}, RoBERTa~\citep{roberta} and BART~\citep{lewis-etal-2020-bart}; two COVID-19-related PLM: BERTweet-covid19~\citep{nguyen-etal-2020-bertweet} and COVID-TWITTER-BERT~\citep{covid-twitter-bert}. BERTweet is RoBERTa-base further trained on 850 million general English tweets and 23 million COVID-19-related tweets. COVID-TWITTER-BERT is BERT-large further trained on 97 million COVID-19-related tweets. All the experiments of NER models are conducted using NCRF++~\citep{yang-zhang-2018-ncrf}. 

\vspace{3mm}
{\bf Training and Test Sets}. To compare the model performance, we perform a train-dev-test splitting of our dataset with a ratio of 70:15:15. Statistics of the train-dev-test splits are presented in Table~\ref{tab:ner_data}. Hyperparameters of the models are default settings from~\cite{yang-etal-2018-design}.

\vspace{3mm}
{\bf Results and Discussion}. Table~\ref{table:ner_benchmark} shows the performance of NER models on the NER test set evaluated by micro-F1. We list both the mean values and standard deviations. We can observe that COVID-TWITTER-BERT achieves the best performance with an overall micro-F1 value of 83.88, outperforming both general domain PLM and classical NER models based on CRF or BiLSTM (and their variants).
Specifically, for the 3 general entity types (\textit{Person}, \textit{Organization} and \textit{Location}), the language models pre-trained on COVID-19-related tweets (COVID-TWITTER-BERT) outperform the best general PLM (RoBERTa-large). The absolute F1 improvements are 3.38, 0.41, and 3.29 for \textit{Person}, \textit{Organization}, and \textit{Location}, respectively. For the 4 medical entity types, COVID-TWITTER-BERT outperforms RoBERTa-large except for the Drug entity type, for which a slightly worse performance is observed.
\begin{table}[t]
\caption{Model performance on METS-CoV-NER dataset. (* means uncased model) }
\begin{center}{
\resizebox{\linewidth}{!}{
\begin{tabular}{ |l | c | c | c | c | c | c | c | c|}

\hline
\rule{0pt}{12pt} 
  {\bf Results (F1 value ± std)}  & {\bf Person} & {\bf Location} & {\bf Organization}  & {\bf Disease} & {\bf Drug} & {\bf Symptom} & {\bf Vaccine} & {\bf Overall}\\
 \hline
CRF  & 64.43±1.59 & 76.37±0.62 & 54.64±2.08  & 73.61±0.44 & 77.34±1.60 & 74.05±0.56 & 84.85±0.82 & 71.58±0.54 \\
\hline
WLSTM & 72.05±0.79 & 79.82±0.61 & 60.79±0.77 & 73.52±1.26 & 79.63±1.36 & 76.72±0.83 & 86.03±0.84 & 75.02±0.36 \\
WLSTM + CCNN & 80.63±0.62 & 81.47±0.89 & 61.30±0.91 & 74.52±0.75 & 80.46±0.28 & 76.63±0.91 & 85.91±1.17 & 76.78±0.29 \\
WLSTM + CLSTM & 81.16±0.29 & 81.37±0.48 & 62.28±1.41 & 74.80±1.49 & 79.50±1.01 & 76.70±0.25 & 85.59±1.46 & 76.91±0.22 \\
WLSTM + CRF & 72.41±0.44 & 79.25±0.59 & 62.39±0.97 & 74.89±1.28 & 79.60±0.71 & \textbf{79.14}±0.51 & 88.72±0.62 & 76.38±0.22 \\
WLSTM + CCNN + CRF & \textbf{81.38}±0.44 & \textbf{82.15}±0.44 & 62.79±0.91 & \textbf{76.12}±0.76 & 80.41±0.58 & 78.12±0.51 & \textbf{89.11}±0.36 & \textbf{78.10}±0.19 \\
WLSTM + CLSTM + CRF & 77.49±1.67 & 81.26±1.19 & \textbf{63.21}±0.93 & 75.61±0.76 & \textbf{81.27}±0.65 & \textbf{79.14}±0.53 & 87.85±0.43 & 77.63±0.40\\
\hline

BERT-base* & 86.99±1.27 & 84.47±0.84 & 71.01±0.52 & 76.18±1.49 & 84.78±1.02 & 80.26±0.43 & 89.83±0.56 & 81.49±0.36  \\
BERT-base & 86.71±0.73 & 84.09±1.67 & 71.90±0.91 & \textbf{76.93}±1.11 & 84.54±1.05 & 79.70±1.22 & 89.48±0.99 & 81.34±0.41  \\
BERT-large* & 87.47±1.22 & 84.43±1.05 & 71.21±0.59 & 76.34±0.96 & 85.53±1.52 & \textbf{81.44}±0.18 & 89.33±1.25 & 81.98±0.30  \\
BERT-large & \textbf{88.25}±0.52 & 84.63±1.38 & 73.30±1.38 & 76.52±1.11 & 86.05±1.06 & 80.12±0.55 & 89.16±1.58 & 82.05±0.24  \\
RoBERTa-base & 85.58±0.73 & 85.46±0.86 & 72.21±1.11 & 76.49±1.63 & 85.38±1.15 & 79.81±0.67 & 89.89±0.28 & 81.43±0.20  \\
RoBERTa-large & 86.79±0.44 & \textbf{85.85}±2.12 & \textbf{73.78}±0.72 & 76.84±0.57 & \textbf{86.79}±0.78 & 81.32±0.67 & \textbf{90.42}±1.12 & \textbf{82.55}±0.27  \\
BART-base & 84.24±0.59 & 82.85±0.71 & 70.60±1.46 & 75.01±1.80 & 83.39±1.03 & 79.03±0.48 & 90.22±1.58 & 80.17±0.46  \\
BART-large & 81.60±4.93 & 80.04±4.74 & 64.66±8.86 & 71.24±1.90 & 80.61±2.90 & 74.27±4.45 & 81.21±6.20 & 75.56±5.04  \\
\hline
BERTweet-covid19-base* & \textbf{91.63}±0.79 & 85.79±0.75 & \textbf{77.07}±0.51 & 77.09±1.61 & 83.57±0.65 & 81.16±0.45 & \textbf{91.16}±1.54 & 83.63±0.36 \\
BERTweet-covid19-base & 91.50±0.81 & \textbf{86.26}±0.97 & 76.47±0.46 & \textbf{77.80}±0.57 & 84.16±1.22 & 80.89±0.52 & 89.98±1.04 & 83.49±0.18  \\
COVID-TWITTER-BERT* & 91.29±0.42 & 85.68±0.92 & 76.27±0.64 & 77.48±0.81 & \textbf{86.35}±0.96 & \textbf{81.85}±0.53 & 90.44±0.94 & \textbf{83.88}±0.20  \\
\hline
 \end{tabular}}}
\label{table:ner_benchmark}
\end{center}
\end{table}
\begin{table}[t]
\centering
\caption{Span F1 of NER models.}
\label{tab:span_f1}
\resizebox{\columnwidth}{!}{%
\begin{tabular}{|l|c|c|c|c|c|c|c|c|}
\hline
\textbf{Model}     & \textbf{Person}     & \textbf{Location}   & \textbf{Organization} & \textbf{Disease}    & \textbf{Drug}       & \textbf{Symptom}    & \textbf{Vaccine}    & \textbf{Overall}    \\ \hline
CRF                & 64.43±1.59          & 76.37±0.62          & 54.64±2.08            & 73.61±0.44          & 77.34±1.60          & 74.05±0.56          & 84.85±0.82          & 74.52±0.77          \\ \hline
WLSTM + CCNN + CRF & 81.38±0.44          & 82.15±0.44          & 62.79±0.91            & 76.12±0.76          & 80.41±0.58          & 78.12±0.51          & 89.11±0.36          & 82.15±0.33          \\ \hline
RoBERTa-large      & 86.79±0.44          & \textbf{85.85}±2.12 & 73.78±0.72            & 76.84±0.57          & \textbf{86.79}±0.78 & 81.32±0.67          & \textbf{90.42}±1.12 & 85.91±0.43          \\ \hline
COVID-TWITTER-BERT & \textbf{90.51}±0.67 & 85.37±0.30          & \textbf{76.31}±0.55   & \textbf{77.14}±1.77 & 86.64±0.65          & \textbf{81.36}±0.38 & 89.71±2.35          & \textbf{86.70}±0.45 \\ \hline
\end{tabular}%
}
\end{table}

\begin{table}[t]
\centering
\caption{Type Acc of NER models.}
\label{tab:type_acc}
\resizebox{\columnwidth}{!}{%
\begin{tabular}{|l|c|c|c|c|c|c|c|c|}
\hline
\textbf{Model}     & \textbf{Person}     & \textbf{Location}   & \textbf{Organization} & \textbf{Disease}    & \textbf{Drug}       & \textbf{Symptom}    & \textbf{Vaccine}    & \textbf{Overall}    \\ \hline
CRF                & 97.25±0.85          & 93.75±0.66          & 93.38±0.57            & \textbf{92.61}±2.07 & 98.00±0.30          & 97.13±0.14          & 97.63±0.53          & 96.06±0.32          \\ \hline
WLSTM + CCNN + CRF & 95.88±0.61          & 93.89±0.15          & 88.08±1.32            & 90.12±1.47          & 97.64±0.90          & 97.42±0.22          & 97.97±0.41          & 95.06±0.20          \\ \hline
RoBERTa-large      & 97.38±0.66          & 95.72±0.89          & 93.87±1.03            & 87.46±0.44          & 98.75±0.86          & \textbf{97.58}±0.34 & \textbf{99.27}±0.42 & 96.18±0.21          \\ \hline
COVID-TWITTER-BERT & \textbf{97.66}±0.34 & \textbf{96.86}±0.83 & \textbf{94.34}±0.77   & 89.38±0.77          & \textbf{99.42}±0.21 & 97.20±0.59          & 97.65±0.57          & \textbf{96.39}±0.16 \\ \hline
\end{tabular}%
}
\end{table}

\begin{figure}[t]
\centering
\begin{minipage}[t]{0.49\textwidth}
\centering
\includegraphics[width=72mm]{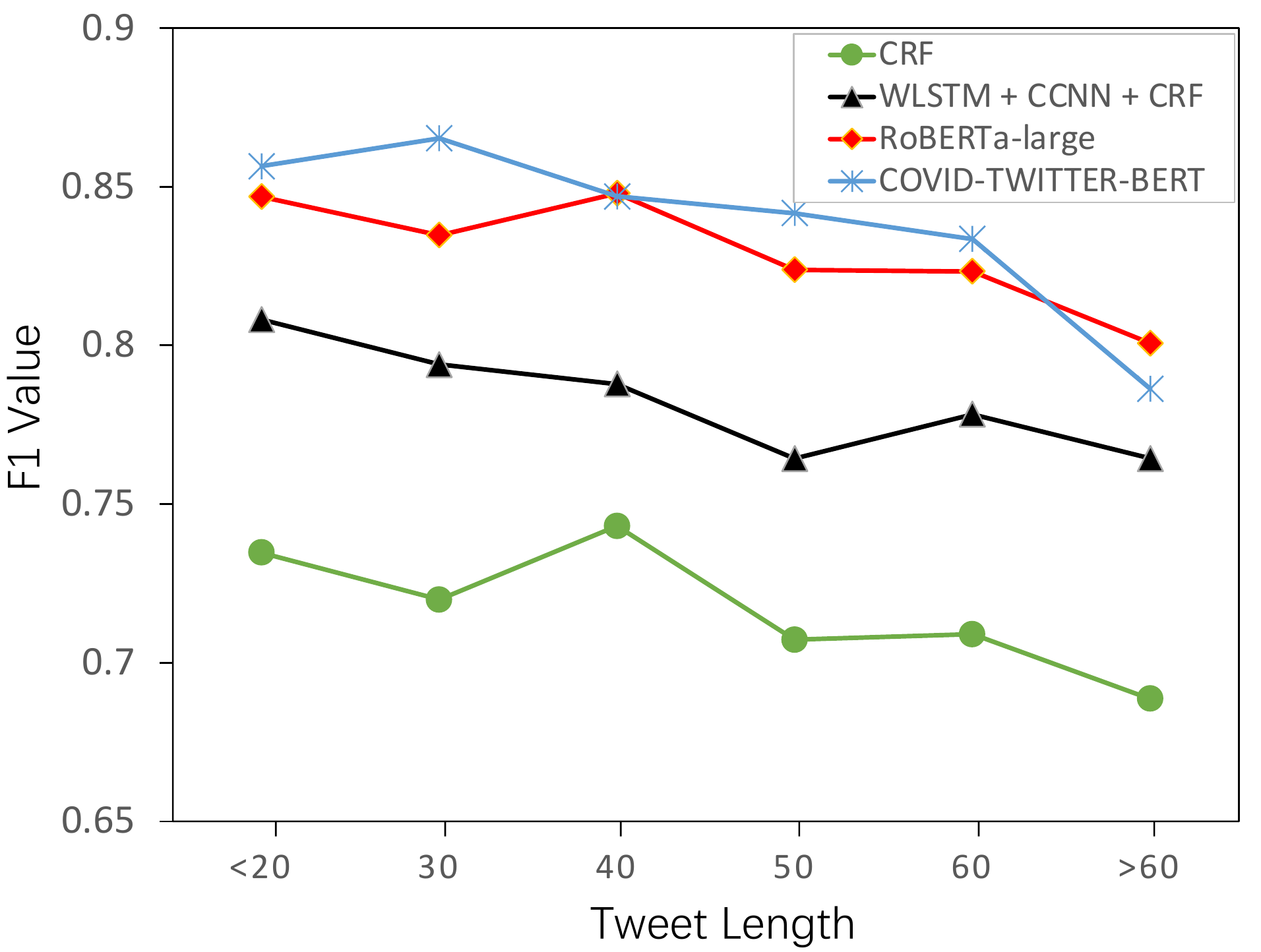}
\caption{F1 values of different NER models against the tweet length.}
\label{Figure: exp_ner_len}
\end{minipage}
\begin{minipage}[t]{0.49\textwidth}
\centering
\includegraphics[width=59mm]{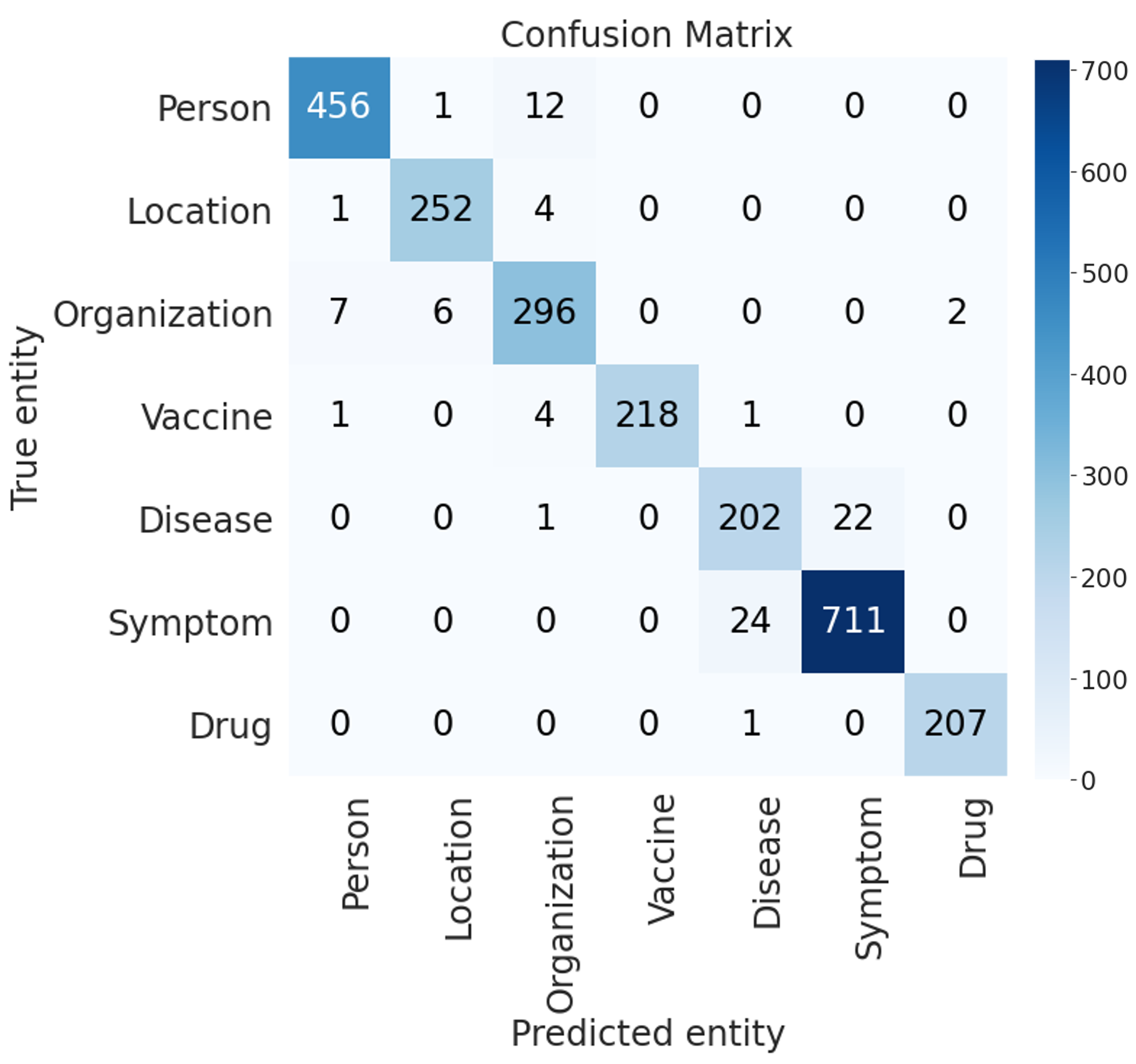}
\caption{The confusion matrix of COVID-TWITTER-BERT on NER test set.}
\label{Figure:exp_ner_confusion}
\end{minipage}
\end{figure}

To compare various models in detail, we select the best models from each branch, i.e., CRF, WLSTM + CCNN + CRF, RoBERTa-large, and COVID-TWITTER-BERT, to represent the state-of-the-art statistical machine learning model, traditional neural network model, general domain PLM, and COVID-19 related PLM, respectively. Following~\cite{liu-etal-2021-lexicon}, we adopt the Span F1 and the Type Accuracy (Type Acc) as the two metrics to evaluate the 4 models. 
Span F1 indicates the correctness of entity spans in NER, while Type Acc refers to the proportion of the predicted entities that have both correct spans and types out of predicted entities with the correct spans.

As shown in Table~\ref{tab:span_f1} and Table~\ref{tab:type_acc}, we observe that COVID-TWITTER-BERT achieves the best overall performance in both metrics, followed by RoBERTa-large.  Specifically, COVID-TWITTER-BERT achieves the best performance on \textit{Person} and \textit{Organization} entities compared to the RoBERTa-large, with 3.72\% and 2.53\% improvement in terms of Span F1, respectively.  For Type Acc, COVID-TWITTER-BERT performs the best on four entity types including \textit{Person, Location, Organization} and \textit{Drug}. These results demonstrate the effectiveness of incremental pre-training of language models on COVID-19-related tweets.
In addition, we also conduct an experiment to investigate the effect of tweet lengths on the model performance. Experimental results are presented in Fig~\ref{Figure: exp_ner_len}. We observe that all models perform better when the tweet length is short (less than 40 tokens). Their performance gradually decreases dealing with longer tweets.
We also visualize the confusion matrix of the best performing model, COVID-TWITTER-BERT, on the test set in Fig~\ref{Figure:exp_ner_confusion}. We find that in most cases, COVID-TWITTER-BERT extracts the entities correctly. However, it tends to get confused when labeling \textit{Symptom} and \textit{Disease} entities. 
This confusion is expected because diseases and symptoms oftentimes have similar expressions and contexts. Besides, we also notice that all models performed worse on \textit{Organization} entities. This problem has also been encountered and discussed in previous Twitter NER dataset papers such as \citep{jiang-EtAl:2022:LREC2}.

Through the above experiments and analysis, we conclude that COVID-TWITTER-BERT can be regarded as a strong baseline on the NER dataset of METS-CoV, and there is still vast room for improvement. Specifically, the F1 value of \textit{Disease} and \textit{Organization} entities is suboptimal. More effective NER models are needed for this challenging dataset. 
\vspace{3mm}
\subsection{Targeted Sentiment Analysis}
{\bf Models}. The TSA models that we benchmark on METS-CoV-TSA could be classified into 4 categories: 1 statistical machine learning model: SVM \citep{vo2015target}; 7 traditional neural network models: ASGCN \citep{zhang-etal-2019-aspect}, LSTM \citep{hochreiter1997long}, TD-LSTM \citep{tang-etal-2016-effective}, MemNet \citep{tang-etal-2016-aspect}, IAN \citep{ijcai2017-568}, MGAN \citep{fan-etal-2018-multi} and TNet-LF~ \citep{li-etal-2018-transformation}; 6  general domain PLM (BERT-base-uncased): 
AEN~\citep{Song2019AttentionalEN}, 
LCF \citep{zeng2019lcf}, BERT-SPC \citep{devlin-etal-2019-bert},  depGCN \citep{zhang-etal-2019-aspect}, kumaGCN \citep{chen2020inducing} and dotGCN \citep{chen-etal-2022-discrete}; and 4 models (BERT-SPC, depGCN , kumaGCN  and dotGCN) with COVID-19 related PLM (COVID-TWITTER-BERT).

\vspace{3mm}
{\bf Train and Test Sets}. The TSA dataset's train-dev-test splitting (with a ratio of 70:15:15) corresponds to that of the NER dataset mentioned above. In this setting, the TSA training dataset is the subset of the NER training dataset, where only the tweets containing targeted entities are kept. A similar procedure is also adopted to obtain the dev and test sets of TSA. The details of TSA split statistics can be found in Table~\ref{tab:tsa_data}. Still, we set the hyperparameters of the models according to the original papers.

\begin{table}[t]
\caption{Model performance on METS-CoV-TSA dataset.
\\($\dagger$ means our implementation. The standard deviation of SVM model is not reported because the prediction of LibSVM\protect\footnotemark[4]
is not affected by random seeds when the dataset splitting is fixed.)}
\begin{center}{
\resizebox{\linewidth}{!}{
\begin{tabular}{| l | c  c| c  c| c  c| c  c | c  c|}
\hline
\multicolumn{1}{|c|}{\multirow{2}[4]{*}{\textbf{Model}}} & \multicolumn{2}{c|}{\textbf{Person}} & \multicolumn{2}{c|}{\textbf{Organization}} & \multicolumn{2}{c|}{\textbf{Drug}} & \multicolumn{2}{c|}{\textbf{Vaccine}} & \multicolumn{2}{c|}{\textbf{Overall}} \bigstrut\\
\cline{2-11}          & \multicolumn{1}{c}{\textbf{Acc}} & \multicolumn{1}{c|}{\textbf{F1}} & \multicolumn{1}{c}{\textbf{Acc}} & \multicolumn{1}{c|}{\textbf{F1}} & \multicolumn{1}{c}{\textbf{Acc}} & \multicolumn{1}{c|}{\textbf{F1}} & \multicolumn{1}{c}{\textbf{Acc}} & \multicolumn{1}{c|}{\textbf{F1}} & \multicolumn{1}{c}{\textbf{Acc}} & \multicolumn{1}{c|}{\textbf{F1}} \bigstrut\\
\hline
SVM \citep{vo2015target}   &50.72  &36.99  &64.57  &42.02  &58.15  &30.17  &70.89  &46.09  &59.53  &38.73  \\
\hline  
  
LSTM \citep{hochreiter1997long}   &58.56±1.79  &50.41±2.48  &61.00±0.95  &45.64±0.47  &56.39±2.18  &41.53±1.92  &65.99±2.41  &40.03±3.62  &60.21±1.53  &49.08±1.58\\
TD-LSTM \citep{tang-etal-2016-effective}   &59.26±0.98  &49.54±1.81  &63.90±2.19  &41.57±2.76  &59.91±1.72  &41.04±2.22  &73.08±0.77  &38.14±2.73  &63.16±0.65  &48.26±1.09\\
MemNet \citep{tang-etal-2016-aspect}  &\textbf{59.79}±1.57  &43.97±3.30  &64.79±2.48  &37.98±1.92  &59.21±1.86  &40.24±1.19  &74.43±1.21  &36.98±2.79  &\textbf{63.73}±0.85  &45.04±1.65 \\
IAN \citep{ijcai2017-568} &52.81±1.72  &32.75±2.65  &\textbf{67.68}±0.45  &36.70±1.94  &59.73±0.45  &25.22±0.10  &\textbf{77.22}±0.00  &30.88±0.00  &62.59±0.55  &34.62±1.77 \\
MGAN \citep{fan-etal-2018-multi} &57.17±2.00  &43.84±4.70  &63.84±2.68  &40.09±1.20  &58.33±2.22  &35.32±5.01  &72.49±0.98  &37.05±5.10  &62.00±1.10  &42.55±3.17\\
TNet-LF \citep{li-etal-2018-transformation}  &58.07±1.19  &\textbf{51.17}±2.10  &63.16±1.52  &\textbf{47.68}±1.59  &\textbf{60.00}±2.00  &\textbf{46.30}±2.51  &68.52±3.15  &\textbf{41.57}±2.17  &61.71±1.01  &\textbf{50.80}±1.22 \\
ASGCN\citep{zhang-etal-2019-aspect}  &58.89±0.63  &42.48±2.61  &63.89±0.84  &39.73±2.62  &58.41±1.90  &31.12±3.91  &72.66±2.02  &38.24±2.95  &62.69±0.23  &41.32±2.22\\
\hline

AEN \citep{Song2019AttentionalEN}  &56.84±2.54  &47.91±4.12  &60.63±4.22  &45.68±3.58  &52.51±3.12  &37.08±4.31  &69.12±4.59  &41.46±3.31  &59.37±2.43  &46.28±3.73 \\
LCF \citep{zeng2019lcf}  &60.29±2.10  &52.43±2.31  &68.58±1.06  &50.15±4.17  &58.42±3.16  &44.01±1.64  &71.14±2.73  &41.75±1.90  &64.27±1.61  &51.29±2.06 \\
BERT-SPC \citep{devlin-etal-2019-bert} $\dagger$   &64.39±0.91  &60.06±1.01  &73.28±0.93  &58.48±1.50  &62.38±1.33  &49.11±2.90  &77.22±0.96  &\textbf{50.28}±5.09  &68.87±0.34  &59.31±1.18\\
depGCN \citep{zhang-etal-2019-aspect}$\dagger$   &\textbf{67.39}±1.16  &62.35±2.08  &\textbf{74.02}±0.71  &\textbf{58.62}±1.06  &\textbf{63.61}±1.07  &49.32±0.74  &77.13±1.54  &47.50±2.73  &\textbf{70.38}±0.40  &59.96±0.77\\
kumaGCN \citep{chen2020inducing}$\dagger$   &66.28±1.33  &61.84±2.46  &72.86±0.59  &58.01±2.04  &\textbf{63.61}±0.77  &49.67±3.14  &76.88±0.73  &50.17±4.56  &69.59±0.73  &59.91±2.48\\
dotGCN \citep{chen-etal-2022-discrete}$\dagger$   &67.06±1.63  &\textbf{62.56}±1.54  &73.33±0.68  &58.34±1.52  &63.35±2.13  &\textbf{50.20}±1.98  &\textbf{77.55}±0.77  &45.98±2.90  &70.09±0.75  &\textbf{60.05}±1.41  \\
\hline
BERT-SPC(COVID-TWITTER-BERT)$\dagger$  &73.72±1.47  &70.25±2.10  &78.53±0.61  &66.24±1.59  &75.07±1.06  &62.67±3.10  &\textbf{79.15}±1.02  &\textbf{61.68}±3.70  &76.29±0.57  &70.03±0.92 \\
depGCN (COVID-TWITTER-BERT) $\dagger$   &\textbf{75.85}±1.22  &\textbf{72.70}±1.78  &\textbf{79.42}±1.25  &\textbf{66.94}±2.66  &\textbf{76.92}±1.49  &\textbf{67.35}±2.45  &77.89±2.14  &59.85±4.45  &\textbf{77.42}±0.77  &\textbf{71.39}±1.65\\
kumaGCN (COVID-TWITTER-BERT) $\dagger$  &74.37±1.39  &71.46±1.43  &78.48±1.46  &64.56±1.61  &76.30±2.38  &62.96±7.41  &78.73±1.50  &58.87±2.63  &76.65±1.20  &70.20±1.95\\
dotGCN (COVID-TWITTER-BERT) $\dagger$   &74.95±1.60  &72.53±1.54  &79.11±0.89  &65.21±2.06  &74.10±1.90  &61.26±2.14  &78.65±1.72  &59.41±2.92  &76.65±0.49  &70.32±0.96 \\
\hline
\end{tabular}}}
\label{Table:absa_benchmark}
\end{center}
\end{table}
\footnotetext[4]{\url{https://github.com/duytinvo/ijcai2015}}
\begin{figure}[t]
\centering
\begin{minipage}[t]{0.49\textwidth}
\centering
\includegraphics[width=65mm]{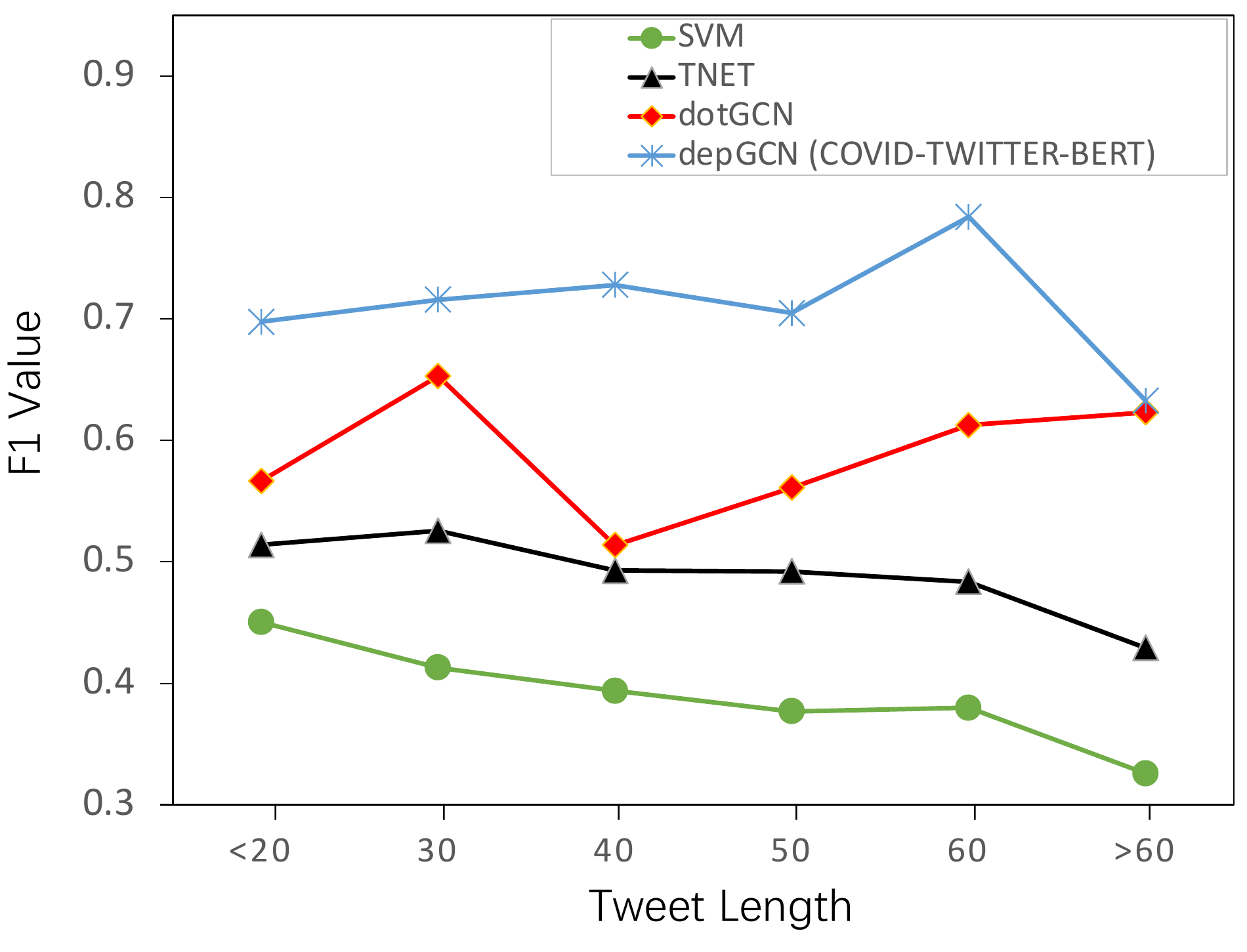}
\caption{F1 values of different TSA models against the tweet length. }
\label{fig:tsa_len}
\end{minipage}
\begin{minipage}[t]{0.49\textwidth}
\centering
\includegraphics[width=65mm]{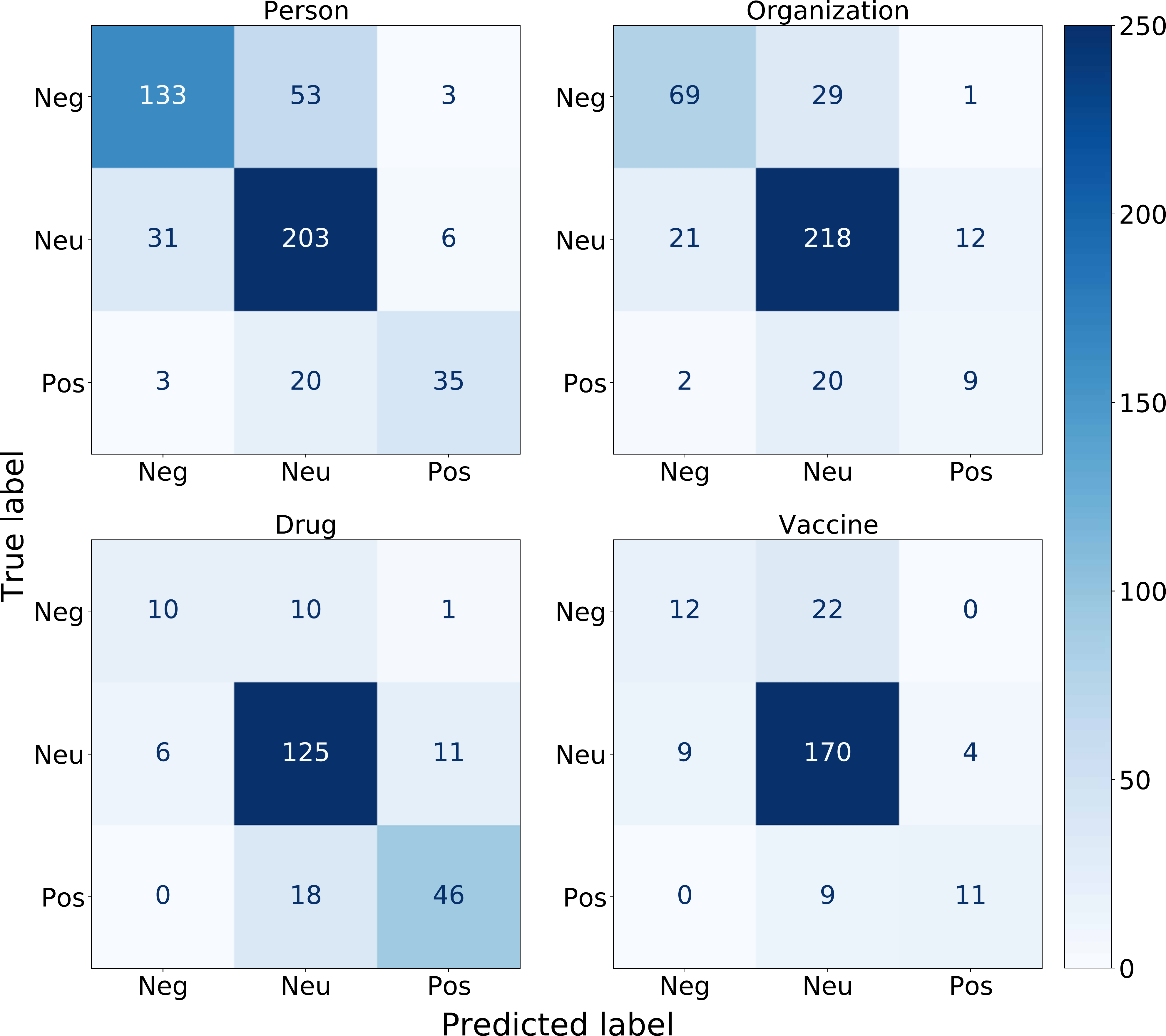}
\caption{The confusion matrix of depGCN (COVID-TWITTER-BERT) on TSA test set. }
\label{fig:tsa_confusion}
\end{minipage}
\end{figure}
\vspace{3mm}
{\bf Results and Discussion}. The experiment results of TSA are listed in Table \ref{Table:absa_benchmark}. We can observe that models incorporating COVID-TWITTER-BERT as feature extractor significantly outperform other types of models. Specifically, depGCN (COVID-TWITTER-BERT) achieves the best performance on the \textit{Person} entities compared to the BERT-based depGCN, with 8.46\% and 10.35\% improvement in terms of ACC and F1, respectively.
For \textit{Organization} entities, depGCN (COVID-TWITTER-BERT) still performs the best, with at least 5.4\% and 8.32\% improvement on ACC and F1, respectively.
For \textit{Drug} entities, depGCN (COVID-TWITTER-BERT) outperforms other models by a large margin and achieves 13.31\% and 18.03\% absolute improvements in terms of ACC and F1 compared with BERT-based depGCN.
For \textit{Vaccine} entities, BERT-SPC (COVID-TWITTER-BERT) outperforms all other models, with at least 1.6\% and 11.4\% improvement in terms of ACC and F1.

To investigate the impacts of tweet lengths on different TSA models, we select the best models from each branch, i.e., SVM, MemNet, depGCN (BERT-base), and depGCN (COVID-TWITTER-BERT). As shown in Fig~\ref{fig:tsa_len}, we observe that the impacts of tweet length varies for different TSA models.
For SVM and TNET, the F1 value gradually decrease when the length of tweets gets longer. For dotGCN, the F1 value fluctuates sharply when the tweet length is between 20 and 40. Afterwards, it increases moderately as the tweet length grows from 40. For depGCN (COVID-TWITTER-BERT), the F1 value remains stable when the tweet length is less than 50 and then increases to 0.8 and finally decreases to about 0.6.

We choose depGCN (COVID-TWITTER-BERT) for an in-depth analysis since it achieves the best overall performance.
Fig~\ref{fig:tsa_confusion} shows the confusion matrix of applying depGCN (COVID-TWITTER-BERT) to the TSA test set. The results show that, for all the targeted entities, most confusion is caused by the misclassification between positive (negative) and neutral.

To summarize, we find that the pre-trained models on COVID-19 tweets, such as COVID-TWITTER-BERT, can be utilized to further improve the performance of previous TSA models on the TSA dataset of METS-CoV. Moreover, the in-depth study shows that the performance of the current best TSA model remains insufficient, and more robust TSA models are needed to distinguish sentiment polarities sufficiently.

\section{Ethics}
\label{ethics}

{\bf General Ethical Conduct}. METS-CoV is the first dataset to include medical entities and targeted sentiments on COVID-19-related tweets. These human-derived data are openly displayed on the Twitter platform and are allowed to be used for research purposes following Twitter's Developer Agreement and Policy.
Following the guidelines, we release only the Tweet IDs but not the original content. Meanwhile, the download script we provide for promoting replicable research could be configured not to store user information. Readers can request desensitized tweets if the tweets can no longer be accessed.
All tweets used in this study were retrieved in an entirely random manner. No new bias should have been introduced except for potentially associated issues with label imbalance. We expect future studies to investigate such issues and provide solutions. We also encourage users to conduct sanity checks to avoid potential bias for specific sub-populations regarding social economics status, cultural groups, etc. Note that we did not filter offensive content in our study because directly removing such words can alter user sentiments expressed in the original tweet. However, we encourage the users to consider doing so in tailored ways for developing fair models.

\vspace{2mm}
{\bf Potential Negative Societal Impacts}. 
Readers should not use the dataset to assess the reputation of public figures or accounts because it was collected and annotated only to develop models for NER and TSA rather than direct causal referential analyses. It is also vital to note that TSA models trained on METS-CoV-TSA should not be used to analyze public attitudes toward people or organizations for non-medical purposes, given that METS-CoV derives from COVID-19-related tweets. Even using such models in a non-COVID-19 setting should be evaluated because public attitudes towards specific topics may change in different settings, and the model may have learned information associated explicitly with COVID-19. Moreover, sentiment analysis for vaccines and drugs can only reveal the user's overall viewpoint and attitude, not their willingness to take drugs or get vaccinated. In addition, readers should be aware that the TSA dataset only reflects user attitudes in the provided contexts and does not explain underlying reasons. Although our work does not directly result in negative social impacts, it is necessary to take appropriate precautions to avoid such impacts.
\section{Limitations}
\label{limitation}
We acknowledge the following limitations: First, METS-CoV has imbalanced entity distribution in that the medical-related tweets are matched using a symptom lexicon to reduce the sparsity of medical entities. To mitigate this problem, we have provided performance evaluation for each entity in the NER benchmark analysis to help readers understand how the models perform on different entities.
Second, we do not filter tweets in MES-CoV-TSA because we want to show the actual distribution of sentiments. This again results in label imbalance, reflected by the high proportion of neutral labels. Readers should be aware of this imbalance when using the dataset.
Third, the TSA annotations unavoidably contain subjectivity. To relieve this problem, we have made strict guidelines, conducted multi-rounds of pre-annotation training, and had annotators work in pairs with third-party validation. 
\section{Conclusion \& Future Work}
In this work, we introduce METS-CoV, the first dataset to include medical entities and targeted sentiments on COVID-19-related tweets. Based on this dataset, we evaluate the performance of both classical and state-of-the-art models for NER and TSA tasks. Results show that existing models can not fully exploit the potential of METS-CoV. 
The METS-CoV dataset is built from a medical research perspective. It fully considers the characteristics of the medical field and can therefore be used to help researchers use natural language processing models to mine valuable medical information from tweets. Much COVID-19 research could leverage this dataset. For example, investigating the public attitudes toward COVID-19 vaccines and drugs, tracking the public's mental status change during different COVID-19 phases, etc. Besides the data, we hope our released annotation guidelines, benchmark models, and source code could facilitate and encourage the curation of more datasets and novel models for medical social medial research.

\begin{ack}
We thank Minghui Li for making the tweet length distribution figure. We appreciate the annotation work of Shixu Lin, Minghui Li, Wanxin Li, Yujie Zhang, Junjie Wang, Subatijiang, and Bingtao Guan. This research received no grant from any funding agency.
\end{ack}

\bibliographystyle{ACM-Reference-Format}
\bibliography{main_and_supp}

\section*{Checklist}

The checklist follows the references.  Please
read the checklist guidelines carefully for information on how to answer these
questions.  For each question, change the default \answerTODO{} to \answerYes{},
\answerNo{}, or \answerNA{}.  You are strongly encouraged to include a {\bf
justification to your answer}, either by referencing the appropriate section of
your paper or providing a brief inline description.  For example:
\begin{itemize}
  \item Did you include the license to the code and datasets? \answerYes{See Section~\ref{gen_inst}.}
  \item Did you include the license to the code and datasets? \answerNo{The code and the data are proprietary.}
  \item Did you include the license to the code and datasets? \answerNA{}
\end{itemize}
Please do not modify the questions and only use the provided macros for your
answers.  Note that the Checklist section does not count towards the page
limit.  In your paper, please delete this instructions block and only keep the
Checklist section heading above along with the questions/answers below.

\begin{enumerate}

\item For all authors...
\begin{enumerate}
  \item Do the main claims made in the abstract and introduction accurately reflect the paper's contributions and scope?
    \answerYes{}
  \item Did you describe the limitations of your work?
    \answerYes{See Section \ref{limitation}}
  \item Did you discuss any potential negative societal impacts of your work?
    \answerYes{See Section \ref{ethics}}
  \item Have you read the ethics review guidelines and ensured that your paper conforms to them?
    \answerYes{}
\end{enumerate}

\item If you are including theoretical results...
\begin{enumerate}
  \item Did you state the full set of assumptions of all theoretical results?
    \answerNA{}
	\item Did you include complete proofs of all theoretical results?
    \answerNA{}
\end{enumerate}

\item If you ran experiments (e.g. for benchmarks)...
\begin{enumerate}
  \item Did you include the code, data, and instructions needed to reproduce the main experimental results (either in the supplemental material or as a URL)?
    \answerYes{All the datasets, benchmarks and code are available at \url{https://github.com/YLab-Open/METS-CoV} }
  \item Did you specify all the training details (e.g., data splits, hyperparameters, how they were chosen)?
    \answerYes{Those details were listed in Section 4. For the hyperparameter selecting, we used the default hyperparameters of the benchmark models. }
	\item Did you report error bars (e.g., with respect to the random seed after running experiments multiple times)?
    \answerYes{Yes, we reported the results based on experiments on 5 different random seeds. Mean ± std were reported in this paper.}
	\item Did you include the total amount of compute and the type of resources used (e.g., type of GPUs, internal cluster, or cloud provider)?
    \answerNo{We didn't include the consumption of resources as we are releasing a new dataset rather than proposing new architecture. }
\end{enumerate}

\item If you are using existing assets (e.g., code, data, models) or curating/releasing new assets...
\begin{enumerate}
  \item If your work uses existing assets, did you cite the creators?
    \answerYes{We used code from several models in our benchmarks, all the sources were properly cited in this paper.}
  \item Did you mention the license of the assets?
    \answerNo{The code we used are all open available, they were used to evaluate model performance in our new dataset. We do not claim any copyright from the code.}
  \item Did you include any new assets either in the supplemental material or as a URL?
    \answerNo{}
  \item Did you discuss whether and how consent was obtained from people whose data you're using/curating?
    \answerYes{See Section \ref{ethics}. This work was conducted on public available data, so this study is waived from the participant's consent. We follow the privacy policy of Twitter platform when sharing this dataset.}
  \item Did you discuss whether the data you are using/curating contains personally identifiable information or offensive content?
    \answerYes{See Section \ref{ethics}}
\end{enumerate}

\item If you used crowdsourcing or conducted research with human subjects...
\begin{enumerate}
  \item Did you include the full text of instructions given to participants and screenshots, if applicable?
    \answerNA{This work was conducted on public available data, it doesn't have participants.}
  \item Did you describe any potential participant risks, with links to Institutional Review Board (IRB) approvals, if applicable?
    \answerNA{This dataset is based on public available tweet text, it doesn't have potential participant risks.}
  \item Did you include the estimated hourly wage paid to participants and the total amount spent on participant compensation?
    \answerNo{This dataset was voluntarily annotated by the authors and members of Prof. Jie Yang's group.}
\end{enumerate}

\end{enumerate}

\end{document}